\documentclass{bmvc2k}

\usepackage{hyperref}
\hypersetup{
    colorlinks=true,
    urlcolor=blue,
    linkcolor=black}

\usepackage{floatrow}
\usepackage[export]{adjustbox}
\usepackage{color}
\usepackage{colortbl}
\usepackage{makecell}
\usepackage{authblk}

\newfloatcommand{capbtabbox}{table}[][\FBwidth]
\usepackage{comment}
\usepackage{amssymb}
\graphicspath{{../figures/}}

\title{TridentAdapt: Learning Domain-invariance via Source-Target Confrontation and Self-induced Cross-domain Augmentation }

\usepackage[symbol]{footmisc}

\footnotetext{$\star$ Equal contribution \hspace{1.2cm} $\dagger$ The majority of the work was done while working at Huawei MRC}

\addauthor{Fengyi Shen}{fengyi.shen@tum.de}{1, 2}
\addauthor{Akhil Gurram}{akhil.gurram@hauwei.com}{$\star$ 2, 3}
\addauthor{Ahmet Faruk Tuna}{a.tuna@gmx.at}{$\star \dagger$ 2}
\addauthor{Onay Urfalioglu}{onay.urfalioglu@gmail.com}{$\dagger$ 2}
\addauthor{Alois Knoll}{knoll@in.tum.de}{1}

\addinstitution{
Department of Informatics,\\
Technical University of Munich, \\
(TUM), 85748 Garching, Germany.} 
\addinstitution{
Huawei Munich Research Center, \\
(MRC), 80992 Munich, Germany.} 
\addinstitution{
Department of Computer Science, \\
Universitat Aut\`onoma de Barcelona,\\
(UAB), 08193 Bellaterra, Spain.} 

\runninghead{Shen, Gurram, Tuna, Urfalioglu, Knoll}{TridentAdapt}

\makeatletter
\DeclareRobustCommand\onedot{\futurelet\@let@token\@onedot}
\def\@onedot{\ifx\@let@token.\else.\null\fi\xspace}

\def\ie{\emph{i.e}\onedot}

\makeatother

\newcommand{\Fig}[1]{Fig. \ref{fig:#1}}
\newcommand{\Eq}[1]{Eq. (\ref{eq:#1})}
\newcommand{\Sect}[1]{Sect. \ref{sec:#1}}

\newcommand{\Tab}[1]{Table \ref{tab:#1}}

\newcommand{\veryshortarrow}[1][3pt]{\mathrel{%
   \hbox{\rule[\dimexpr\fontdimen22\textfont2-.2pt\relax]{#1}{.4pt}}%
   \mkern-4mu\hbox{\usefont{U}{lasy}{m}{n}\symbol{41}}}}

\definecolor{maroon}{cmyk}{0,0.87,0.68,0.32}
\definecolor{Gray}{gray}{0.9}
\definecolor{LightCyan}{rgb}{0.88,1,1}
\definecolor{Red}{rgb}{1,0,0}
\definecolor{Green}{rgb}{0,1,0}
\definecolor{Blue}{rgb}{0,0,1}
\definecolor{Yellow}{rgb}{1,1,0}
\definecolor{Orange}{rgb}{1,0.5,0}

\begin{document}

\maketitle
\vspace{-3mm}

\begin{abstract}
Due to the difficulty of obtaining ground-truth labels, learning from virtual-world datasets is of great interest for real-world applications like semantic segmentation. From domain adaptation perspective, the key challenge is to learn domain-agnostic representation of the inputs in order to benefit from virtual data. 

In this paper, we propose a novel trident-like architecture that enforces a shared feature encoder to satisfy confrontational source and target constraints simultaneously, thus learning a domain-invariant feature space. Moreover, we also introduce a novel training pipeline 
enabling self-induced cross-domain data augmentation during the forward pass. This contributes to a further reduction of the domain gap. Combined with a self-training process, we obtain state-of-the-art results on benchmark datasets (e.g. GTA5 or Synthia to Cityscapes adaptation). Code and pre-trained models are available at {\href{https://github.com/HMRC-AEL/TridentAdapt}{https://github.com/HMRC-AEL/TridentAdapt}}

\end{abstract}

\section{Introduction}
\label{sec:intro}
Deep neural networks have shown tremendous potential in dealing with computer vision challenges such as semantic segmentation~\cite{chen2017deeplab, chen2017rethinking, long2015fully, yuan2019object}, image recognition~\cite{simonyan2014very, szegedy2015going, he2016deep, huang2017densely, dosovitskiy2020image}, etc. Semantic segmentation, a fundamental building block in autonomous driving systems, refers to the task of classifying each pixel in an image belonging to a certain semantic class. Unfortunately, the creation of datasets with pixel-wise labels is notably a laborious and high-cost procedure. 

Benefiting from the development of modern computer graphics technology such as game simulators~\cite{richter2016playing,ros2016synthia,Dosovitskiy17}, the appearances of real-world objects can be well imitated via virtual \pagebreak imagery. Moreover, fine-grained semantic image labels can be acquired in large-scale for free from those virtual environments, which opens a new research direction for computer vision applications. However, due to lighting and texture differences between objects in virtual and real-world images, direct knowledge transfer from virtual to real is limited. Well-trained models on virtual source datasets often experience drastic performance drop when they are applied on real-world target domain where labels are missing. 

\begin{figure}
\centering
\includegraphics[width=0.77\columnwidth]{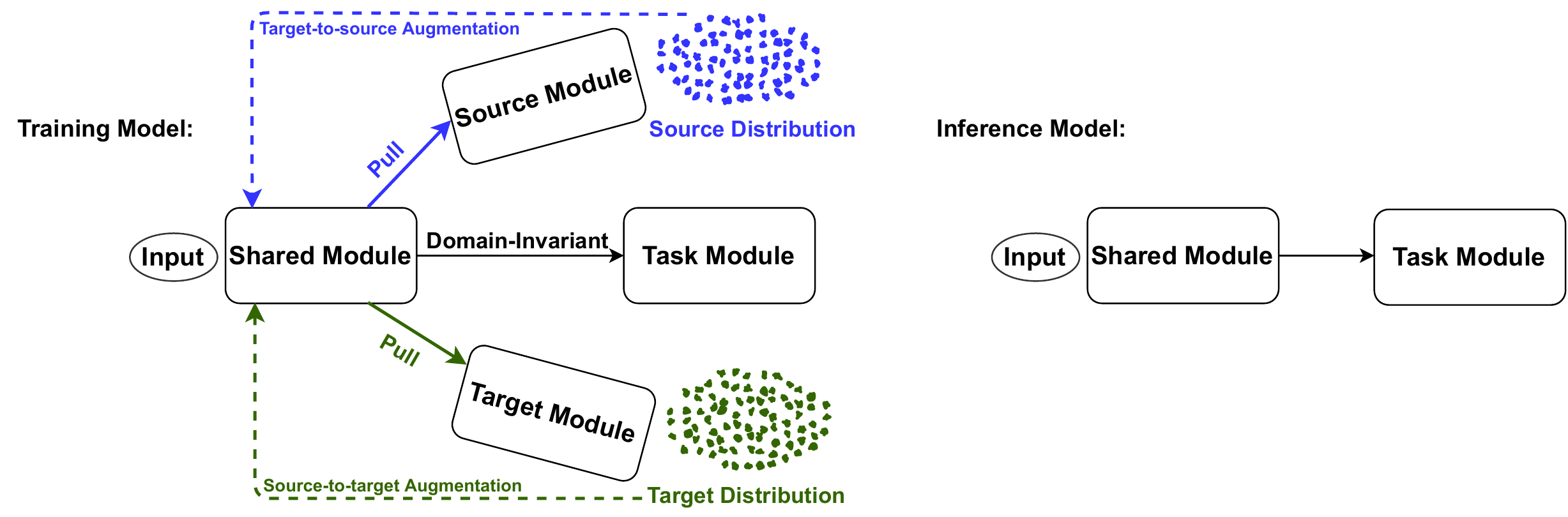}
\caption{ {\bf Algorithmic Overview}. For any given input data (source or target domain), we leverage the source and target modules to put confrontational constraints to the shared module simultaneously, forcing it to produce a domain invariant representation of the feature map. The novel augmented views induced from source and target modules are then fed back to the shared module to further bridge the domain gap. The learning of feature domain-invariance will boost the performance of the task module on target domain data. Only the shared and task modules will be adopted for inference. }
\label{fig:model_description}
\vspace{-3.5mm}
\end{figure}


After showing promising results in pixel-level image style transfer, image-to-image translation~\cite{huang2018multimodal,liu2017unsupervised,zhu2017unpaired} models have brought much attention to unsupervised domain adaption. A pioneer research is conducted in~\cite{hoffman2018cycada}, which makes use of target-like images translated from source domain together with their labels to provide guidance when learning target domain segmentation. Despite the visually pleasing target-like translations, important features get lost during image translation, because a pure image-to-image translation model is not built for conveying semantic information, additionally it does not guarantee a perfect mapping from source to target domain. Thus, the performance gain for semantic segmentation is limited on target domain data.

Other interesting approaches such as curriculum domain adaptation~\cite{zhang2017curriculum}, depth-aware domain adaptation~\cite{vu2019dada} and frequency domain adaptation~\cite{yang2020fda} also obtain state-of-the-art results for domain adaptive semantic segmentation. Unfortunately, adaptable knowledge is still not fully explored and transferred from source to target domain, and hence remains an open topic.

In this work, we provide a new perspective for solving the domain adaptive semantic segmentation problem. We hypothesize that a shared feature map which can help produce a source as well as a target domain output simultaneously, should be domain-invariant. To this end, as \Fig{model_description} illustrates, we design our TridentAdapt framework such that the shared module is forced to satisfy both source and target constraints, resulting in learning domain-invariant representations from input data. With this design, the framework further benefits from its back-fed cross-domain augmented views that are self-induced by source and target modules during training to bridge the domain gap.  

We summarize our contributions as follows:
\begin{itemize}
\item We put forward an intuitive yet effective trident-like architecture in which source and target distributions adopt a confrontational stance, compelling the shared encoder to produce feature maps that are indistinguishable in terms of domain. 
\item We incorporate cross-domain augmented views which are self-induced from our framework into the training pipeline. Thus, we introduce semantic consistency losses on feature level and extra segmentation losses on output level, which not only encourage semantic information to be conveyed into the generators for better output quality, but also boost the learning of domain-invariance for the encoder and segmentation network. 
\item Our trained models demonstrate superior results over state-of-the-art methods on challenging benchmark datasets.
\end{itemize}

\section{Related Work}
\noindent{\bf Input Level Adaptation.} 
Since image-to-image translation models~\cite{zhu2017unpaired,liu2017unsupervised,huang2018multimodal} based on GANs~\cite{goodfellow2014generative,arjovsky2017wasserstein, mao2017least,jolicoeur2018relativistic,karras2019style} can be trained following an unsupervised manner, in~\cite{hoffman2018cycada,yue2019domain} target-like images translated from source domain are involved to train a target segmentation model for cross-domain improvement. \cite{chen2019crdoco} employs for source and target domains separate segmentation networks, to which source data and their target-like translations are fed, respectively, to force the segmentation maps to be consistent. In~\cite{toldo2020unsupervised}, other than adversarial feature alignment, all possible cross-domain outputs based on CycleGAN training pipeline are taken into consideration to reduce the domain gap. \cite{musto2020semantically} improves source-to-target translation by incorporating SPADE layers~\cite{park2019semantic}. In another work~\cite{kim2020learning}, semantic segmentation is learned by receiving target-like translations together with the stylized source images carrying various texture changes to prevent the segmentation network from overfitting on one specific source texture. SG-GAN~\cite{li2018semantic} employs a gradient-sensitive loss and a semantic-aware discriminator to improve structural contents after image translation. However, the above approaches seek to build separate networks for image translation and semantic segmentation purposes, where domain transfer modules and segmentation encoder do not place sufficient constraints on each other. Therefore, the potential of image translation is not fully explored to support domain adaptive semantic segmentation. Our proposed method tackles this by sharing the learned semantic knowledge across all networks and looping back cross-domain outputs to collaboratively learn a reinforced domain-agnostic feature space. 

\noindent{\bf Output Level Adaptation.} Aligning segmentation outputs between source and target domains is considered an effective way to narrow the domain gap. Adversarial learning is adopted in~\cite{tsai2018learning,vu2019advent} by connecting source and target segmentation outputs to a discriminator which learns structured output space. \cite{yang2020label} proposes to refine segmentation by connecting a image reconstruction network to the output label maps.  
Self-training~\cite{chapelle2009semi,zhu2005semi} such as pseudo-label generation, which enhances the confidence level of knowledge learned from source data, has become a widely adopted concept in domain adaptive semantic segmentation. In~\cite{choi2019self,du2019ssf,kim2020learning,li2019bidirectional,pan2020unsupervised,zou2018unsupervised,zou2019confidence}, self-training is conducted to acquire pseudo-labels which further enhance segmentation performance on target domain.


\noindent{\bf Feature Level Adaptation.}
In semantic segmentation models, deeper layer features often convey rich semantic information. Therefore, exploring a domain-invariant feature space, which can be shared by both source and target input data, will theoretically bring significant effect on minimizing cross-domain discrepancy. However, this has always been a key challenge in domain adaptive semantic segmentation. Inspired by adversarial learning, a feature discriminator is introduced in~\cite{hoffman2018cycada} to force the encoder of segmentation network to extract similarly distributed feature maps from both domains. Taking one step forward, ~\cite{du2019ssf} makes use of the downsampled pseudo-labels to design a semantic-wise separable feature discriminator which allows class level adversarial learning to improve the domain invariance of deep features. In~\cite{sankaranarayanan2018learning} and ~\cite{zhu2018penalizing}, the segmentation encoder is connected by a shared generator to produce domain transferred outputs alternately to reduce domain gap on feature level. However, the generated images are not utilized to further interact with the segmentation network for mutual improvement.  
DCAN~\cite{wu2018dcan} seeks to align channel-wise statistics of bottleneck features between source and target domains via AdaIn~\cite{huang2017arbitrary}.
Inspired by~\cite{huang2018multimodal}, ~\cite{chang2019all} is trained to split input data into domain-specific texture and domain-invariant structure, and the learned structural features are used to train the segmentation network. In our framework, however, we implicitly put source and target distribution in confrontation to help search for a domain-invariant feature space, obtaining better performance than existing approaches. 

\begin{figure}[]
\centering
\includegraphics[width=0.8\columnwidth]{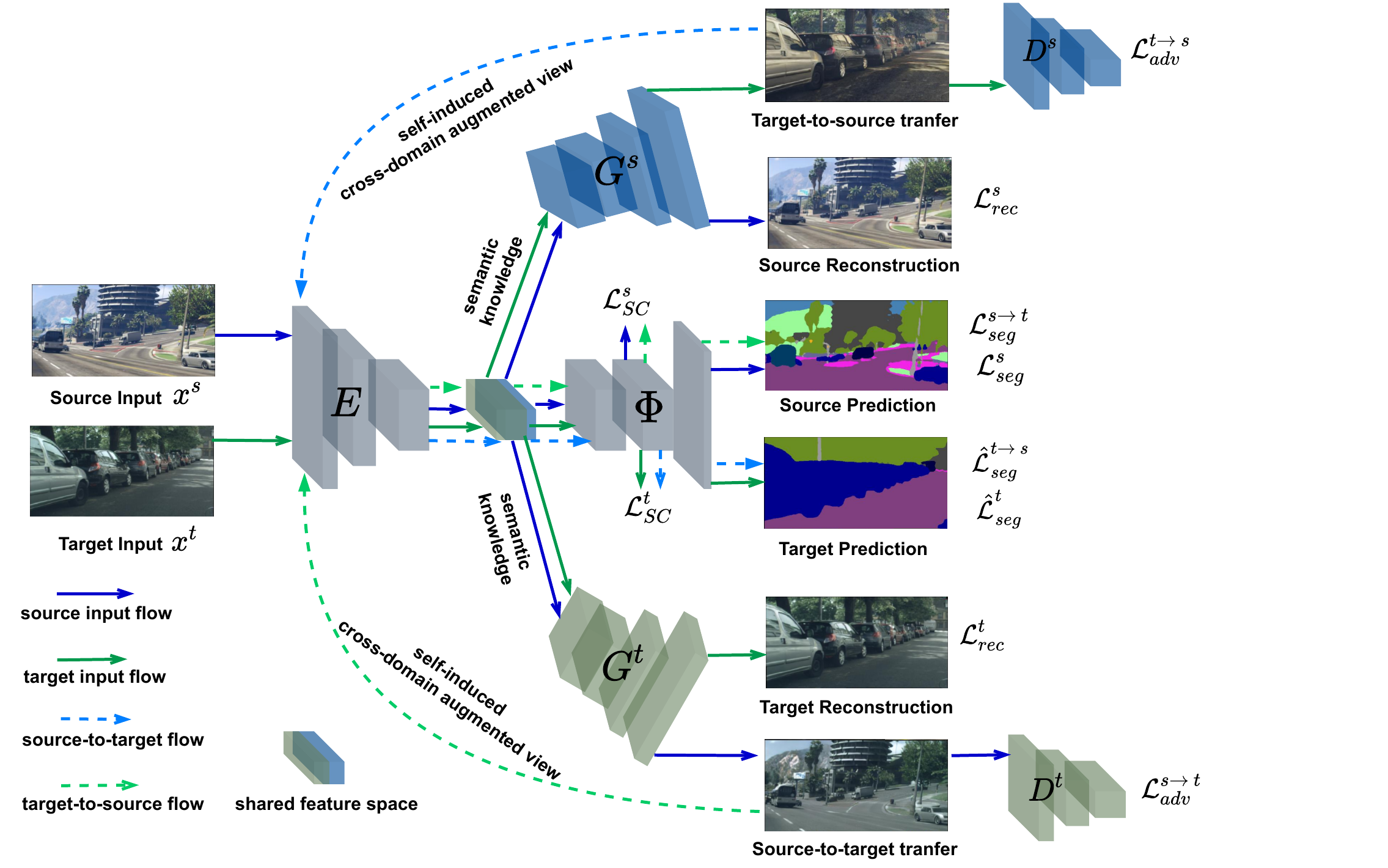}
\caption{A pictorial overview of our proposed TridentAdapt framework in detail.}
\label{fig:model_diagram}
\vspace{-3.5mm}
\end{figure}


\vspace{-3.5mm}
\section{Proposed Method}
\label{sec:proposed_method}
In this section, we introduce TridentAdapt framework for domain adaptive semantic segmentation. Let $\{X^s, Y^s\}$ and $X^t$ denote the source and target domain datasets respectively, where $x^s \in X^s$ stands for a source training RGB image with corresponding source label map $y^s \in Y^s$, and $x^t \in X^t$ stands for a target training image whose label $y^t \in Y^t$ is missing. The goal is to train a model that is able to predict correct per-pixel label for $X^t$ by the assistance of $\{X^s, Y^s\}$. As depicted in \Fig{model_description}, we achieve this by leveraging a confrontation between source and target distributions to learn a domain-invariant feature space for the segmentation network ${\Phi}$. The learning is reinforced by introducing self-induced cross-domain augmented data into a backward loop, bridging the domain gap further. Detailed pictorial description is provided in \Fig{model_diagram}.

\subsection{Source-target confrontation}
\label{sec:source-target}
To achieve our design purpose, effective data distribution modelling for both domains is a key step. Although classical image-to-image translation approaches~\cite{huang2018multimodal, liu2017unsupervised, zhu2017unpaired} exhibit some clues on how image distribution can be modelled using GANs, yet they focus more on altering image appearances for style transfer, ignoring the semantic level information. Therefore, for modelling source and target distributions we propose semantic-aware generators, to which semantic knowledge from the encoder is incorporated. As shown in \Fig{model_diagram}, TridentAdapt training pipeline contains mainly four modules: shared encoder ${E}$ initialized by weights of ImageNet~\cite{deng2009imagenet} pretrained backbone network, source generator ${G^{s}}$, target generator ${G^{t}}$ and segmentation network ${\Phi}$. 

If ${E}$ receives an image ${{x}^{s}}$ from source domain dataset during training and outputs a feature map ${E(x^{s})}$, based on which ${G^{s}}$ will produce a source reconstruction image ${G^{s}({E(x^{s})})}$ measured by L$1$-loss ${\mathcal{L}^{s}_{rec}}$ in \Eq{img_rec_s}. Note that for ${\mathcal{L}^{s}_{rec}}$, pixels belonging to image edges are prioritized in order to compensate the omitted pixels due to the max pooling operations in the encoder.  
\begin{align}
    \label{eq:img_rec_s}
    {\mathcal{L}^{s}_{rec}({E}, {G^{s}}; {X}^{s})} = \mathbb{E}_{{x}^{s} \sim {X}^{s}} [||{\Omega^{s} \odot(G^{s}({E(x^{s}))}} - {x}^{s})||_{1}]
\end{align}

$\Omega^{s}$ $\in \mathcal{R}^{\tiny c^{im} \times \tiny h \times \tiny w}$, is a weight matrix where $\omega^{s}_{ijk}= 
\begin{cases}
    1+\eta^{s},& \text{if } x^{s}_{ijk}\in edges \\
    1,              & \text{otherwise}
\end{cases}$\\ 
(${i}$ stands for the channel index and ${j}, {k}$ the spatial indices). $\odot$ denotes element-wise multiplication. Empirically we set $\eta^{s}$ to $0.5$.
Here the $edges$ are computed based on a sobel operator, where ${h}$, ${w}$ and $c^{im}$ are height, width and the number of channels of an image $x^s$ respectively.

In parallel, ${G^{t}}$ takes the same feature map ${E(x^{s})}$ and generates a source-to-target transferred image ${G^{t}({E(x^{s})})}$ following target distribution with help of the target discriminator ${D^{t}}$, computing adversarial loss ${\mathcal{L}^{s\,\veryshortarrow\,t}_{adv}}$ based on LSGAN~\cite{mao2017least} for target domain,
\begin{align}
    \label{eq:trans_adv_s2t}
    \mathcal{L}^{s\,\veryshortarrow\,t}_{adv} (E, G^{t}, D^{t}; {X}^{t}, {X}^{s}) = \mathbb{E}_{ {x}^{t}\sim {X}^{t}} [(D^{t}({x}^{t}))^2] + \mathbb{E}_{{x}^{s} \sim {X}^{s}} [(1 - D^{t}( G^{t}( {E(x^{s})})))^2]
\end{align}

\noindent Here the domain-agnostic feature is learned by the confrontational constraints coming from intra-domain reconstruction (pulling towards source) and cross-domain transfer (pulling towards target) simultaneously.

At the same time, in the middle path of TridentAdapt, ${\Phi}$ is the task module which takes encoded feature maps for semantic segmentation. By default we denote the output of ${\Phi}$ as an upsampled probability map which has been processed through softmax operation. As ground truth is known for each ${{x}^{s}}$, during training we minimize the cross entropy loss ${\mathcal{L}^{s}_{seg}}$ using source prediction ${\Phi({E(x^{s})})}$ under the supervision of ${{y}^{s}}$:

\begin{align}
    \label{eq:source_seg}
    &{\mathcal{L}^{s}_{seg}(E,\Phi; {X}^{s}, {Y}^{s})} = -\mathbb{E}_{({x}^{s}, {y}^{s})\sim ({X}^{s}, {Y}^{s})} \sum_{h,w,c} {y}^{s}_{(h,w,c)} \log( \Phi({E(x^{s})}))_{(h,w,c)}
\end{align}
\noindent 

Here $c$ is the number of semantic classes. Since ${E}$ also serves as feature extractor for ${\Phi}$, semantic information will thus be incorporated into ${G^{t}}$ to make it semantic-aware while generating ${G^{t}({E(x^{s})})}$. 


On the other way round, if ${E}$ receives an image ${{x}^{t}}$ from target domain dataset (unlabelled), following a symmetric data flow, the confrontational constraints on feature map ${E(x^{t})}$ can be computed similar to \Eq{img_rec_s} and \Eq{trans_adv_s2t} to obtain ${\mathcal{L}^{t}_{rec}}$ and ${\mathcal{L}^{t\,\veryshortarrow\,s}_{adv}}$ respectively.


In the above procedure, based on input data domain, ${G^{s}}$ and ${G^{t}}$ are updated according to a role-switching mechanism, e.g., during training ${G^{s}}$ is always an image decoder from perspective of $x^{s}$ but is adopted as image translator for $x^{t}$. By switching the role of the generators according to the switch of input data, our framework guarantees that ${G^{s}}$ only produces source images, ${G^{t}}$ purely outputting target images regardless of the input data domains. In this way, for example, when ${G^{s}}$ is used to reconstruct $x^{s}$, it learns how `real' source data distribution should be reflected on its output, and this knowledge is expected to provide weak guidance to fine-tune itself when it takes the next role as image translator for $x^{t}$ to generate a `fake' source image ${G^{s}({E(x^{t})})}$, 
contributing to reinforced target to source transfer.

\subsection{Self-induced Cross-domain Augmentation}
\label{sec:self_induced}
The shared trident-like design of our framework allows us to utilize its own outputs to enhance the learning of domain-invariance in a self-served manner. We introduce a backward loop where the self-induced cross-domain augmented views ${G^{t}({E(x^{s})})}$ and ${G^{s}({E(x^{t})})}$ are fed to ${E}$. As those views never appear in training set but they resemble cross-domain data distributions, such that ${E}$ receives a broader coverage of input data. In this way, the augmented views that are self-induced on-the-fly during each iteration can provide a smooth transition to bridge the domain gap between source and target domains. To enable this, we introduce semantic consistency(SC) loss ${\mathcal{L}_{SC}}$ to force the encoder to consider each input and its cross-domain version semantically identical. 

Specifically, we take the semantic features from an intermediate layer of the segmentation network ${\Phi}$, which is divided into 2 blocks ${\Phi} = \{\Phi_f, \Phi_s\}$. Now for each domain we obtain one feature pair, \ie,
${{\Phi_f}({E(x^{s})})}$ and ${{\Phi_f}(E(G^{t}({E(x^{s})})))}$ for source, ${{\Phi_f}({E(x^{t})})}$ and ${{\Phi_f}(E(G^{s}({E(x^{t})})))}$ for target. 
Minimizing the feature distance of each pair, we compute:
\begin{align}
    \label{eq:SC}
    \mathcal{L}^{s}_{SC} ( &E, {\Phi_f}, G^{s}; {X}^{s}) = \mathbb{E}_{{x}^{s} \sim {X}^{s}}[|| {\Phi_f}(E(G^{t}({E(x^{s})}))) - {\Phi_f}({E(x^{s})})||_{1}]\\
    \mathcal{L}^{t}_{SC}( &E, {\Phi_f}, G^{t}; {X}^{t}) = \mathbb{E}_{{x}^{t} \sim {X}^{t}}[|| {\Phi_f}(E(G^{s}({E(x^{t})}))) - {\Phi_f}({E(x^{t})})||_{1}]
\end{align}

In addition, for each input and its generator outputs, we add a VGG-based perceptual loss~\cite{johnson2016perceptual} which is widely adopted for training image-to-image translation models, maintaining feature-level structure between input and output images. We get ${\mathcal{L}^{s_{rec}, s}_{percep}}$, ${\mathcal{L}^{s\veryshortarrow t,s}_{percep}}$, ${\mathcal{L}^{t_{rec}, t}_{percep}}$ and ${\mathcal{L}^{t\veryshortarrow s,t}_{percep}}$ based on L1-metric following exactly~\cite{johnson2016perceptual}.

Since ${E}$ is shared across ${G^{s}}$, ${G^{t}}$ and ${\Phi}$, with ${\mathcal{L}^{s}_{SC}}$ and ${\mathcal{L}^{t}_{SC}}$ we also improve ${G^{s}}$ and ${G^{t}}$ to carry higher-level semantic information. We choose ${{\Phi_f}}$ instead of ${E}$ bottleneck features, as it contains richer task-specific information. 


\noindent\textbf{Augmented Semantic Segmentation}.
For further fine-tuning the segmentation head using source-to-target augmented view, we let $G^{t}({E(x^{s})})$ share the same ground truth ${{y}^{s}}$, computing another supervised segmentation loss ${\mathcal{L}^{s\veryshortarrow\,t}_{seg}}$ similar to \Eq{source_seg}. Thus, ${\Phi}$ gradually adapts itself to images which display target domain characteristics.

\noindent\textbf{Self-training}.
Although ground-truths for target domain are missing, the above training steps can still provide large support for learning domain-invariance to improve segmentation performance on target domain. Therefore, we first train our framework for a warming-up stage which is referred to as `stg1'. Thereafter for target domain dataset we perform pseudo-labelling~\cite{li2019bidirectional} offline based on predictions of ${\Phi}$ (class-wise median probability as threshold) to acquire a set of pseudo-labels ${\Hat{{Y}}^{t}}$, with which we go for a self-training phase 'stg2'.

Upon the acquisition of ${{\Hat{{Y}}^{t}}}$, we are able to enhance segmentation for target domain data simply by adding loss ${\Hat{\mathcal{L}}^{t}_{seg}}$ during each training iteration, 
\begin{align}
    \label{eq:target_seg}
    &{\Hat{\mathcal{L}}^{t}_{seg} ( E, \Phi; {X}^{t}, \Hat{{Y}}^{t})} =  -\mathbb{E}_{({x}^{t}, \Hat{{y}}^{t})\sim({X}^{t}, \Hat{{Y}}^{t})} \sum_{h,w,c}\Hat{{y}}^{t}_{(h,w,c)}\log(\Phi({E(x^{t})}))_{(h,w,c)}
\end{align}

Similarly, we also fine-tuned the segmentation head further by computing $\hat{\mathcal{L}}_{seg}^{t\,\veryshortarrow\,s}$ using the self-induced target-to-source view $G^{s}({E(x^{t})})$ which enjoys the same pseudo-label.

\subsection{Full objective}


We denote $\mathbb{N}$ as the set of names for all above losses, summing up all of which formulates our full objective for training TridentAdapt: 
\begin{align}
    \label{eq:full_loss}
    \mathcal{L}_{\mathrm{TridentAdapt}} = \sum_{i\in\mathbb{N}} \lambda_{i}\mathcal{L}_{i}
\end{align}

\section{Experiments and Discussion}
\label{sec:experiments}
In this section, we provide our experimental setups and report our results on benchmark datasets for comparison with state-of-the-art methods.

\subsection{Datasets and Implementation Details}
For target domain we adopt Cityscapes dataset~\cite{cordts2016cityscapes} containing 2975 annotated street scene images with $2048\times 1024$ resolution (annotations excluded during training) and 500 images for validation, and we consider for source domain the GTA5 dataset~\cite{richter2016playing} consisting of 24,966 annotated images with $1914\times 1052$ resolution taken from game engine, as well as SYNTHIA-RAND-CITYSCAPES dataset~\cite{ros2016synthia} consisting of 9,400 images of $1280\times 760$ resolution with fine-grained segmentation labels.

We implement TridentAdapt with Pytorch~\cite{NEURIPS2019_9015} on an NVIDIA Quadro RTX 8000 with 48 GB memory, of which around 37 GB will be taken when running our training scripts. For all experiments we use pretrained {\tt ResNet-101}~\cite{he2016deep} as backbone feature extractor to initialize encoder ${E}$ and adopt {\tt Deeplab-V2}~\cite{chen2017deeplab} for segmentation network ${\Phi}$. During training, we first resize source input images to $1280\times 720$ resolution and target images to $1024\times 512$ resolution but take $512\times 256$ random crops for both domains in each training iteration. We train our framework using batch size $4$ and set the max training iteration number to $2.5\times 10^{5}$ (Though the model performs equally well even before $2\times 10^{5}$ iterations). We use the SGD~\cite{robbins1951stochastic} optimizer with a default learning rate of $2.5\times 10^{-4}$ for ${E}$ and $\Phi$,  and Adam~\cite{adam:2015} optimizers for $G^{s}, G^{t}$ with default learning rate $1.0\times 10^{-3}$ but $1.0\times 10^{-4}$ for $D^{s}, D^{t}$. Polynomial decay policy is applied to all learning rates. We set momentum to $0.9$ and $0.99$. For evaluation on target validation set we upsample the segmentation predictions to the full resolution of Cityscapes~\cite{cordts2016cityscapes} dataset. 

Notice that for the first $3500$ iterations in `stage {1}' we detach feature maps from ${E}$ before passing to $G^{s}$ and $G^{t}$, aiming to let them warm up and not to deteriorate the pretrained encoder ${E}$ in the initial training stage. Starting from the $5000$th iteration we take source-to-target transferred images to compute segmentation loss and all cross-domain augmented images to compute semantic consistency losses as they look realistic enough. 

We adopt multi-scale discriminator architecture for $D^{s}$ and $D^{t}$ following~\cite{huang2018multimodal}.
During training, we weight the losses with different hyperparameters, here empirically we set $\lambda^{s}_{rec}$ =  $\lambda^{t}_{rec}$ = 1,  $\lambda^{s\,\veryshortarrow\,t}_{adv}$ = $\lambda^{t\,\veryshortarrow\,s}_{adv}$ = 0.1,  $\lambda^s_{sc}$ = $\lambda^t_{sc}$ = 0.1,  $\lambda^{s}_{seg}$ = $\lambda^{s\,\veryshortarrow\,t}_{seg}$ = 1, $\Hat{\lambda}^{t}_{seg}$ = $\Hat{\lambda}^{t\,\veryshortarrow\,s}_{seg}$ = 0.75, $\lambda^{s_{rec}, s}_{percep}$ = $\lambda^{t_{rec}, t}_{percep}$ = 0.5, $\lambda^{s\veryshortarrow t, s}_{percep}$ = $\lambda^{t\veryshortarrow s, t}_{percep}$ = 0.25.

\subsection{Comparison with State of the Art}

\begin{figure}[htb!]
\centering
\includegraphics[width=0.83\columnwidth,clip=true,trim=0 23 0 0]{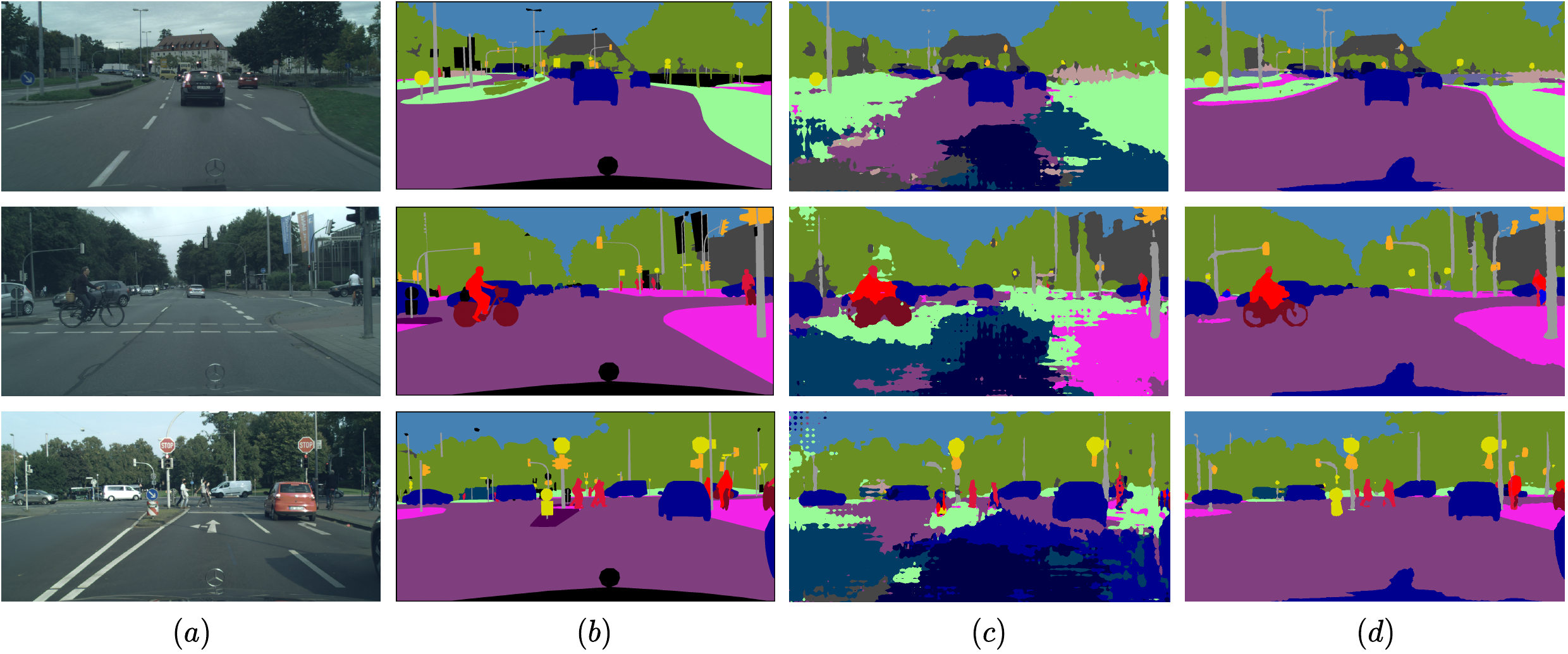}
\caption{ Qualitative results of GTA5-to-Cityscapes adaptation on Cityscapes validation set. Columns from left to right are: target domain inputs; ground-truth labels; segmentation predictions of source-only model;  segmentation predictions of TridentAdapt.}
\label{fig:seg_output}
\vspace{-3.5mm}
\end{figure}

Our TridentAdapt approach shows leading performance among the state-of-the-art methods on GTA5-to-Cityscapes adaptation presented in \Tab{gta5tocity}, achieving $53.3$ mIoU. Moreover, our approach outperforms other methods by considerable margins on many challenging classes (e.g., pole, traffic light, trafic sign, person, rider, motorcycle, etc.). \Fig{seg_output} visually presents examples of the segmentation results by  TridentAdapt. On Synthia-to-Cityscapes adaptation in \Tab{synthiatocity}, we also achieved state-of-the-art results, being superior to other methods in segmenting cars, buses, etc. Our framework is compatible with other data augmentation techniques such as ~\cite{yun2019cutmix} and ~\cite{olsson2021classmix} to achieve better results (See Supplementary Sec.4).


\subsection{Ablation study}

In this section, we investigate the effectiveness of TridentAdapt components. In rows 1-8 of \Tab{ablationstudy}, we add each component at a time to show the performance of segmentation task on GTA5-to-Cityscapes adaptation. The 1st row stands for training with source only data, whereas the 2nd row is our baseline which is taken from~\cite{tsai2018learning}. The 3rd row shows that training with $G^{s}\,\&\,G^{t}$ will bring confrontational constraints (described in \Sect{source-target}) to ${E}$ to learn feature domain-invariance, yielding a performance increase of $0.7$ over the baseline, and adding VGG perceptual losses in row 4 gives $0.3$ improvement. In the 5th row, involving self-induced cross-domain augmentations (described in \Sect{self_induced}) to compute ${\mathcal{L}^s_{sc}} \, \& \, {\mathcal{L}^t_{sc}}$ leads to a performance boost to $44.6$ mIoU, which suggests that domain gap can be bridged by our self-induced augmentations, widening input coverage in domain level. By introducing $\mathcal{L}^{s\,\veryshortarrow\,t}_{seg}$ to the previous setup (5th row) for fine-tuning $\phi$ brings another performance gain of $1.9$ mIoU as can be seen in 6th row. Note that the result seen in 6th row, which totally relies on adversarial learning for domain adaptation, already outperforms some existing self-training based approaches~\cite{choi2019self, du2019ssf, pan2020unsupervised, zou2018unsupervised}. Finally, adding pseudo-labelling by computing $\Hat{\mathcal{L}}^{t}_{seg}, \, \Hat{\mathcal{L}}^{t\,\veryshortarrow\,s}_{seg}$ leads to the best performance in 8th row, which increases mIoU from $46.5$ to $53.3$. This is conceivable since ${E}$ is shared across $G^{s}$, $G^{t}$ and ${\Phi}$, any improvement on ${E}$ means other modules can benefit as well, thus enhancing the learning objectives among each other. Row 9 indicates that training without looping back our self-induced augmentations will lead to a performance drop from 53.3 to 47.6. Overall, this analysis shows how all the considered components are relevant in our proposal.

\begin{table}[htb!]
    \centering
    \small
    \setlength{\tabcolsep}{2.5pt}
    \fontsize{6}{8.2}\selectfont
    \begin{tabular}{c|ccccccccccccccccccc|c}
    \hline Method & \rotatebox{90}{road} & \rotatebox{90}{sdwk} & \rotatebox{90}{bldng} & \rotatebox{90}{wall} & \rotatebox{90}{fence} & \rotatebox{90}{pole} & \rotatebox{90}{light} & \rotatebox{90}{sign} & \rotatebox{90}{veg} & \rotatebox{90}{trrn} & \rotatebox{90}{sky} & \rotatebox{90}{psn} & \rotatebox{90}{rider} & \rotatebox{90}{car} & \rotatebox{90}{truck} & \rotatebox{90}{bus} & \rotatebox{90}{train} & \rotatebox{90}{moto} & \rotatebox{90}{bike} & mIoU  \\ 
    \hline
    \\[-1.5em]
    Source-only~\cite{tsai2018learning}   & 75.8 & 16.8 & 77.2& 12.5& 21.0& 25.5& 30.1& 20.1& 81.3& 24.6& 70.3& 53.8& 26.4& 49.9&  17.2&  25.9& 6.5& 25.3& 36.0&36.6 \\
    \\[-1.5em]
    AdaptSeg~\cite{tsai2018learning}  &  86.5&  36.0& 79.9& 23.4& 23.3& 23.9& 35.2& 14.8& 83.4& 33.3& 75.6& 58.5& 27.6& 73.7&  32.5&  35.4& 3.9& 30.1& 28.1& 42.4\\
    \\[-1.5em]
    ADVENT~\cite{vu2019advent}  &  89.4&  33.1& 81.0& 26.6& 26.8& 27.2& 33.5& 24.7& 83.9& 36.7& 78.8& 58.7& 30.5& 84.8&  \underline{38.5}&  44.5& 1.7& 31.6& 32.4& 45.4\\
    \\[-1.5em]
    SSF-DAN~\cite{du2019ssf}&  90.3&  38.9& 81.7& 24.8& 22.9& 30.5& 37.0& 21.2& 84.8& 38.8& 76.9& 58.8& 30.7& 85.7&  30.6&  38.1& 5.9& 28.3& 36.9& 45.4\\
    \\[-1.5em]
    CRST~\cite{zou2019confidence} &  91.0&  {\bf55.4}& 80.0& 33.7& 21.4& \underline{37.3}& 32.9& 24.5& 85.0& 34.1& 80.8& 57.7& 24.6& 84.1&  27.8&  30.1& {\bf26.9}& 26.0& 42.3& 47.1\\
    \\[-1.5em]
    PyCDA~\cite{lian2019constructing} &  90.5&  36.3& 84.4& 32.4& 28.7& 34.6& 36.4& 31.5& {\bf86.8}& 37.9& 78.5& \underline{62.3}& 21.5& 85.6&  27.9&  34.8& 18.0& 22.9& {\bf49.3} & 47.4\\
    \\[-1.5em]
    BDL~\cite{li2019bidirectional} &  91.0&  44.7& 84.2& 34.6& 27.6& 30.2& 36.0& 36.0& 85.0& \underline{43.6}& 83.0& 58.6& 31.6& 83.3&  35.3&  \underline{49.7}& 3.3& 28.8& 35.6& 48.5\\
    \\[-1.5em]
    IntroDA~\cite{pan2020unsupervised} &  90.6&  36.1& 82.6& 29.5& 21.3& 27.6& 31.4& 23.1& 85.2& 39.3& 80.2& 59.3& 29.4& \underline{86.4}&  33.6&  {\bf53.9}& 0.0& 32.7& 37.6&46.3 \\
    \\[-1.5em]
    TIR~\cite{kim2020learning} &  {\bf92.9}&  \underline{55.0}& \underline{85.3}& 34.2& \underline{31.1}& 34.9& 40.7& 34.0& 85.2& 40.1& {\bf87.1}& 61.0& 31.1& 82.5&  32.3&  42.9& 0.3& \underline{36.4}& \underline{46.1}& 50.2\\
    \\[-1.5em]
    SAIT~\cite{musto2020semantically}&  91.2&  43.3& 85.2& \underline{38.6}& 25.9& 34.7& \underline{41.3}& \underline{41.0}& 85.5& {\bf46.0}& \underline{86.5}& 61.7& \underline{33.8}& 85.5&  34.4&  48.7& 0.0& 36.1& 37.8& \underline{50.4}\\
    \hline
    \\[-1.5em]
    TridentAdapt &  \underline{91.3}&  51.5& {\bf86.4}& {\bf38.8}& {\bf36.4}& {\bf42.3}& {\bf45.4}& {\bf42.0}& \underline{86.6}& 36.4& 84.3& {\bf67.7}& {\bf42.8}& {\bf89.1}& {\bf41.7}& 38.2 &\underline{20.6} & {\bf40.3}& 30.7&{\bf53.3} \\\hline
\end{tabular}
\caption{GTA5-to-Cityscapes adaptation results. we compare our model performance with state-of-the-art methods which are trained with {\tt ResNet-101}~\cite{he2016deep} and {\tt Deeplab-V2}~\cite{chen2017deeplab} based models. In all the tables of \Sect{experiments}, bold stands for {\bf best}, and underline for \underline{second-best}.}
\label{tab:gta5tocity}
\vspace{-3.5mm}
\end{table}

\begin{table}[htb!]
  \centering
\small
\setlength{\tabcolsep}{2.5pt}
\fontsize{6}{8.2}\selectfont
\begin{tabular}{c|cccccccccccccccc|cc}
\hline Method & \rotatebox{90}{road} & \rotatebox{90}{sdwk} & \rotatebox{90}{bldng} & \rotatebox{90}{$\textrm{wall}^\star$} & \rotatebox{90}{$\textrm{fence}^\star$} & \rotatebox{90}{$\textrm{pole}^\star$} & \rotatebox{90}{light} & \rotatebox{90}{sign} & \rotatebox{90}{veg} & \rotatebox{90}{sky} & \rotatebox{90}{psn} & \rotatebox{90}{rider} & \rotatebox{90}{car} &  \rotatebox{90}{bus} &  \rotatebox{90}{mcycl} & \rotatebox{90}{bcycl} & mIoU & $\textrm{mIoU}^\star$  \\ \hline
    \\[-1.5em]
    Source-only~\cite{tsai2018learning}  &  55.6&  23.8& 74.6& -& -& -& 6.1& 12.1& 74.8& 79.0& 55.3& 19.1& 39.6& 23.3&  13.7&  25.0& -& 38.6 \\
    \\[-1.5em]
    AdaptSeg~\cite{tsai2018learning}  &  84.3&  42.7& 77.5& -& -& -& 4.7& 7.0& 77.9& 82.5& 54.3& 21.0&  72.3&  32.2& 18.9& 32.3& -& 46.7\\
    \\[-1.5em]
    ADVENT~\cite{vu2019advent}  &  85.6&  42.2& 79.7& 8.7& 0.4&  25.9& 5.4& 8.1& 80.4& 84.1& 57.9& 23.8&  73.3&  36.4& 14.2& 33.0& 41.2& 48.0\\
    \\[-1.5em]
    SSF-DAN~\cite{du2019ssf}&  84.6&  41.7& 80.8& -& -& -& 11.5&  14.7& 80.8& \underline{85.3}& 57.5& 21.6&  82.0&  36.0& 19.3& 34.5& -& 50.0\\
    \\[-1.5em]
    CRST~\cite{zou2019confidence} &  67.7&  32.2& 73.9& \underline{10.7}& {\bf1.6}& {\bf37.4}& \underline{22.2}& \underline{31.2}& 80.8& 80.5& \underline{60.8}& \underline{29.1}&  82.8&  25.0& 19.4& 45.3& 43.8& 50.1\\
    \\[-1.5em]
    PyCDA~\cite{lian2019constructing} &  75.5&  30.9& {\bf83.3}& {\bf20.8}& 0.7& 32.7& {\bf27.3}& {\bf33.5}& {\bf84.7}& 85.0& {\bf64.1}& 25.4&  \underline{85.0}&  45.2& 21.2& 32.0& \underline{46.7}& 53.3\\
    \\[-1.3em]
    BDL~\cite{li2019bidirectional} &  86.0&  46.7&80.3& -& -& -& 14.1& 11.6& 79.2& 81.3& 54.1& 27.9&  73.7&  42.2& \underline{25.7}& \underline{45.3}& -& 51.4\\
    \\[-1.5em]
    IntroDA~\cite{pan2020unsupervised} &  84.3& 37.7& 79.5& 5.3& 0.4& 24.9& 9.2& 8.4& 80.0& 84.1& 57.2& 23.0&  78.0&  38.1& 20.3& 36.5& 41.7& 48.9\\
    \\[-1.5em]
    TIR~\cite{kim2020learning} &  {\bf92.6}&  {\bf53.2}& 79.2&-&-&-& 1.6& 7.5& 78.6& 84.4& 52.6& 20.0& 82.1& 34.8& 14.6&  39.4&  -& 49.3 \\
    \\[-1.5em]
    SAIT~\cite{musto2020semantically} &  87.7&  49.7&\underline{81.6}&-& -& -& 19.3& 18.5& \underline{81.1}& 83.7& 58.7& {\bf31.8}&  73.3&  \underline{47.9}& {\bf37.1}& {\bf45.7}& -& {\bf55.1}\\
\hline
   \\[-1.5em]
    TridentAdapt & \underline{89.5}&  \underline{51.9}& 79.1& 7.3& \underline{1.1}& \underline{34.3}& 15.2& 25.8&  80.4& {\bf88.0}& 57.3&  19.2&  {\bf87.5}& {\bf52.2}&18.6 & 42.1&{\bf46.8}&\underline{54.4} \\\hline
\end{tabular}
\caption{Synthia-to-Cityscapes adaptation results. mIoU, ${\textrm{mIoU}^\star}$ refer to {16}-class and {13}-class experiment settings, respectively.}
\label{tab:synthiatocity}
\vspace{-3.5mm}
\end{table}




\begin{figure}
\begin{floatrow}
\capbtabbox{%
    \fontsize{8}{8.2}\selectfont
    
    \begin{tabular}{l|c}
        \hline
        configuration                             &   mIoU    \\ \hline
        \\[-0.75em]
        1. ${X}^{s}, {Y}^{s}$~\cite{tsai2018learning} &   36.6    \\ \hline
        \\[-0.75em]
        2. $ + {X}^{t}$~\cite{tsai2018learning}     &   42.4    \\
        \\[-0.75em]
        3. $ + G^{s} \,\& \,G^{t}$                       &   43.1    \\
        \\[-0.75em]
        4. $ + {\mathcal{L}_{percep}}$      &   43.4    \\
        \\[-0.75em]
        5. $ + {\mathcal{L}^s_{sc}} \, \& \, {\mathcal{L}^t_{sc}}$      &   44.6    \\
        \\[-0.75em]
        6. $ + {\mathcal{L}^{s\,\veryshortarrow\,t}_{seg}}$ (\textbf{stg1})          &   46.5   \\
        \\[-0.75em]
        7. $ + \Hat{\mathcal{L}}^{t}_{seg}$        &   51.8    \\
        \\[-0.75em]
        8. $ + \Hat{\mathcal{L}}^{t\,\veryshortarrow\,s}_{seg}$ (\textbf{stg2})    &   {\bf 53.3}    \\ \hline
        \\[-0.75em]
        9. \makecell{ $  - {\mathcal{L}^s_{sc}} \, \, \& \, \, {\mathcal{L}^t_{sc}} $   \\ $ - \Hat{\mathcal{L}}^{t\,\veryshortarrow\,s}_{seg}  - {\mathcal{L}^{s\,\veryshortarrow\,t}_{seg}}$  \\ }
          &   47.6    \\ \hline
    \end{tabular}
}{%
    \caption{Ablation Study on GTA5-to-Cityscapes adaptation}
    \label{tab:ablationstudy}
    \vspace{-3.5mm}
    }
\ffigbox[9.0cm]{%
    \centering
    \includegraphics[width=1.0\columnwidth,clip=true,trim=10 0 20 5,left]{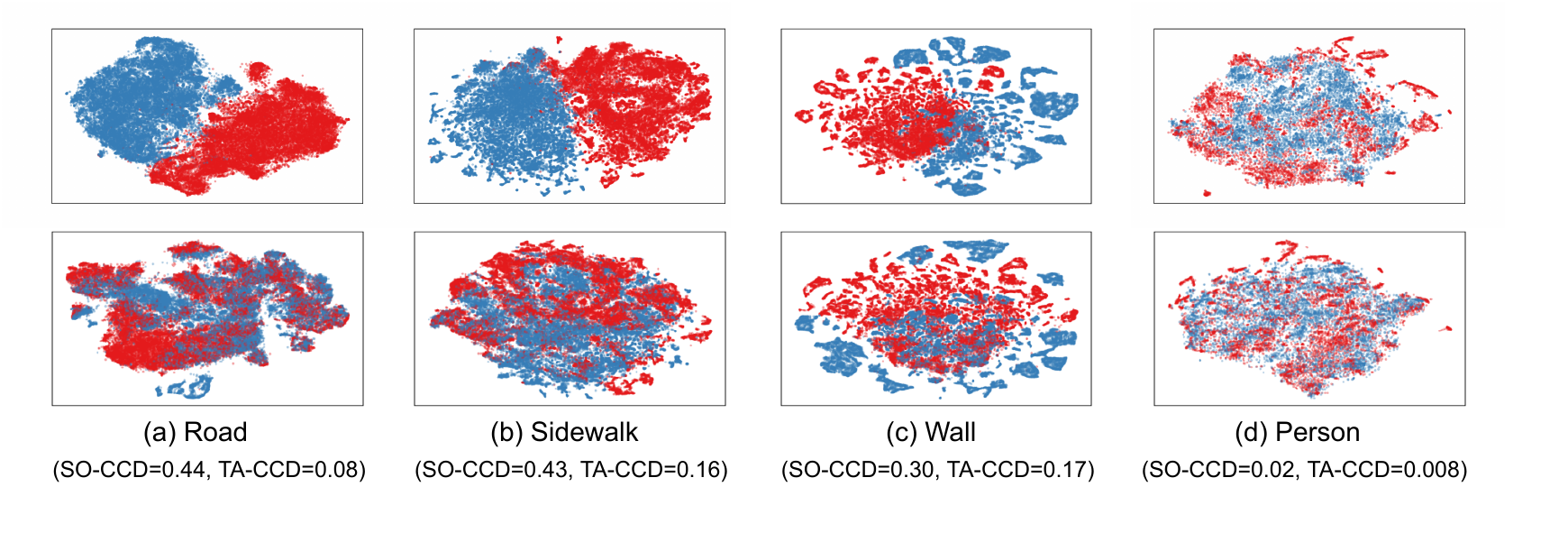}
}{%
    \caption{Class-wise t-SNE \cite{vanDerMaaten2008} visualization: GTA5 features (red) vs. Cityscapes features (blue). From top to bottom: source only (SO-CCD) before adaptation and TridentAdapt model (TA-CCD) after adaptation.
    }
    \label{fig:tsne_analysis}
    \vspace{-3.5mm}
}
\end{floatrow}
\end{figure}

\subsection{Learning Domain-invariance} 
We show empirically that the shared encoder $E$ reduces the domain gap at feature-level by 
visualizing the ${\Phi_f}$ feature-map distribution (described in \Sect{source-target}) and projecting the high dimensional features class-wise on a 2D-space using the t-SNE \cite{vanDerMaaten2008} algorithm. We compare the class-wise feature distributions before adaptation (source-only model) and after adaptation (TridentAdapt). In addition to the visualization we compute for each class the Cluster-Center-Distance (CCD) bewteen source and target feature vectors.  
\Fig{tsne_analysis} illustrates the analysis on GTA5-to-Cityscapes adaptation for four different classes (road, sidewalk, wall, and person). From road, sidewalk and wall classes we can clearly see the source and target domain features of the source-only model are building two different clusters, which are perfectly separable. In comparison, however, we can see that our TridentAdapt model is able to narrow the feature-distribution gap for these classes and can reduce the CCD drastically. For person class, although the features of source-only model are not as fully separable as in previous examples, nevertheless by comparing the CCD we can also confirm that TridentAdapt reduces the domain gap at feature level. 
This analysis implies that the proposed TridentAdapt approach enforces the shared encoder to produce domain-invariant features, such that the cross-domain discrepancy between source and target input data is reduced. 


\begin{figure}
\begin{floatrow}
\capbtabbox{%
    \fontsize{8}{8.2}\selectfont
    \centering
    \begin{tabular}{c|cc}
    \hline
    \\[-0.75em]
    Translation Model   &       mIoU        &   $\Delta$    \\ \hline
    \\[-0.75em]
    Source-only (ours)     &       31.5        &       -       \\\hline
    \\[-0.75em]
    CycleGAN (T2S `validate')            &       35.8        &       +4.3    \\
    \\[-0.75em]
    TridentAdapt (T2S `validate')       &       {\bf 38.6}        &       +7.1    \\ \hline
    \\[-0.75em]
    CycleGAN (S2T `train')     &       39.3        &       +7.8        \\
    \\[-0.75em]
    TridentAdapt (S2T `train')       &       {\bf 44.5}        &       +13.0 \\ \hline    
    \end{tabular}
}{%
    \caption{Quantitative comparison of image translation results for semantic segmentation.}%
    \label{tab:translationstudy}
    \vspace{-3.5mm}
    }
\ffigbox[6.5cm]{%
  \includegraphics[width=\columnwidth]{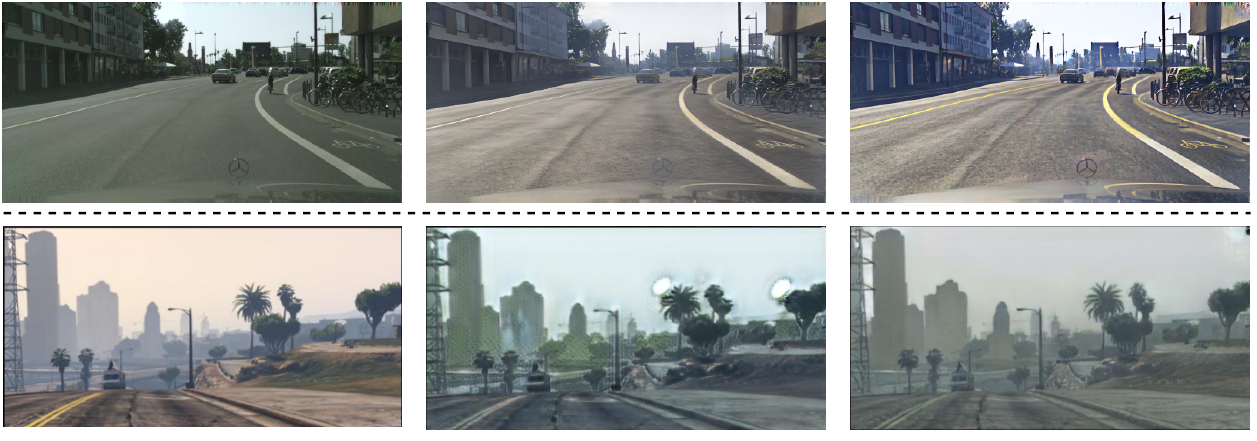}%
}{%
    \caption{Visual comparison of image translation results. From left to right: input, CycleGAN and TridentAdapt output.}%
    \label{fig:compare_cyc_trident}
    \vspace{-3.5mm}
}
\end{floatrow}
\end{figure}
\subsection{Modelling source and target distributions}


We demonstrate experimentally the visibility of employing our proposed semantic-aware generators to model source and target distributions, therefore placing strong domain-specific constraints to the shared encoder in achieving domain-invariance and providing substantial support for bridging domain gap when introduced into a backward loop. We first evaluate the target-to-source image translation performance on Cityscapes validation set. Specifically, we compare CycleGAN~\cite{zhu2017unpaired} and TridentAdapt translated images by passing them to a source-only (GTA5) segmentation model, which is, for fair comparison, trained using our configurations (e.g., batch size, crop size, augmentations) as in TridentAdapt. We observe in \Tab{translationstudy} that our source-like images achieve a $7.1$ gain in mIoU over non-translated Cityscapes validation images, outperforming CycleGAN result also by a large margin.
To evaluate our target generator $G^{t}$, we train a segmentation model sorely on our source-to-target translated GTA5 images, computed mIoU of the trained model on Cityscapes validation set, and compared with the result of CycleGAN~\cite{zhu2017unpaired}. Interestingly, this improves the source-only model result by $13.0$ in mIoU and outperforms CycleGAN~\cite{zhu2017unpaired} by $5.2$.
\Fig{compare_cyc_trident} visually reveals that TridentAdapt better preserves semantic contents (e.g., distant vegetation and vehicles) during translation. Therefore we conclude that generators of TridentAdapt are effective for modelling domain data distributions, which is beneficial to the subsequent tasks for learning domain-invariance.

\section{Conclusion}

We propose TridentAdapt, a trident-like architecture for domain adaptation including a source module and a target module which simultaneously imposes confrontational constraints on the shared feature encoder. We present a novel framework which produces self-induced cross-domain augmentations during the forward pass to further reduce domain gap. Experimental results show SOTA semantic segmentation performance on target domain data. 

\bibliography{tridentadapt}
\end{document}